%% file: main2020.tex
\documentclass{article}

\usepackage{microtype}
\usepackage{graphicx}
\usepackage{subfigure}
\usepackage{booktabs} 
\usepackage{natbib}
\usepackage{xcolor}
\usepackage{amsmath}
\usepackage{amsfonts}
\usepackage{mathtools}
\newcommand{\E}{{\mathbb E}} 

\newcommand{\EI}{{\text{EI}}}
\newcommand{\EIpu}{{\text{EIpu}}}

\input{./macros.tex} 
\usepackage{hyperref}

\usepackage[accepted]{icml2020}

\icmltitlerunning{Cost-aware Bayesian Optimization}

\begin{document}

\twocolumn[
\icmltitle{Cost-aware Bayesian Optimization}

\begin{icmlauthorlist}
\icmlauthor{Eric Hans Lee}{a}
\icmlauthor{Valerio Perrone}{b}
\icmlauthor{C\'edric Archambeau}{b}
\icmlauthor{Matthias Seeger}{b}
\end{icmlauthorlist}

\icmlaffiliation{a}{Cornell University. Work done while interning at Amazon, Berlin, Germany.}
\icmlaffiliation{b}{Amazon, Berlin, Germany}

\icmlcorrespondingauthor{Eric Hans Lee}{ehl59@cornell.edu}
\icmlcorrespondingauthor{Valerio Perrone}{vperrone@amazon.com}
\icmlcorrespondingauthor{C\'edric Archambeau}{cedrica@amazon.com}
\icmlcorrespondingauthor{Matthias Seeger}{matthis@amazon.com}


\icmlsetsymbol{equal}{*}

\icmlkeywords{Machine Learning, ICML}

\vskip 0.3in
]



\printAffiliationsAndNotice{}  

\begin{abstract}
Bayesian optimization (BO) is a class of global optimization algorithms, suitable for minimizing an expensive objective 
function in as few function evaluations as possible. While BO budgets are typically given in iterations, this implicitly measures convergence in terms of iteration count and assumes each evaluation has identical cost. In practice, evaluation costs may vary in different regions of the search space. For example, the cost of neural network training increases quadratically with layer size, which is a typical hyperparameter. \textit{Cost-aware BO} measures convergence with alternative cost metrics such as time, energy, or money, for which vanilla BO methods are unsuited. We introduce Cost Apportioned BO (CArBO), which attempts to minimize an objective function in as little cost as possible. 
CArBO combines a cost-effective initial design with a cost-cooled optimization phase which depreciates a learned cost model as iterations proceed. On a set of 20 black-box function optimization problems we show that, given the same cost budget, CArBO finds significantly better hyperparameter configurations than competing methods.
\end{abstract}

\section{Introduction}
Consider minimizing a black-box function $f(\xb): \Omega \rightarrow \mathbb{R}$ over a convex set $\Omega \subset \mathbb{R}^d$ whose analytical form and gradients are unavailable, and that can only be queried through potentially noisy evaluations. Bayesian optimization (BO) is a well-established class of methods to address this problem, and has been applied with success to a range of tasks, from hyperparameter optimization (HPO) to simulation tuning~\cite{Jones1998, snoek2012practical, Shahriari2016, ju2017designing, frazier2018}. BO is \textit{sample efficient}, taking fewer steps to converge than competing global optimization methods. Evaluations $f(\xb_1),\ldots,f(\xb_n)$ are used to model $f$, typically with a Gaussian process (GP) \citep{rasmussen2006gaussian}. An acquisition function, implicitly defined by the GP, balances exploration and exploitation to determine the next evaluation. A popular choice is the Expected Improvement (EI) \citep{Mockus1978}, defined as $\EI(\xb) \coloneqq \E \Big[  \max \big(y^* - f(\xb)), 0\big) \Big]$, which is the expected reduction in the objective of an evaluation with respect to the current minimum $ y^* \coloneqq f(\xb_{min})$.

\begin{figure*}[th] 
    \centering
    \includegraphics[width=\textwidth]{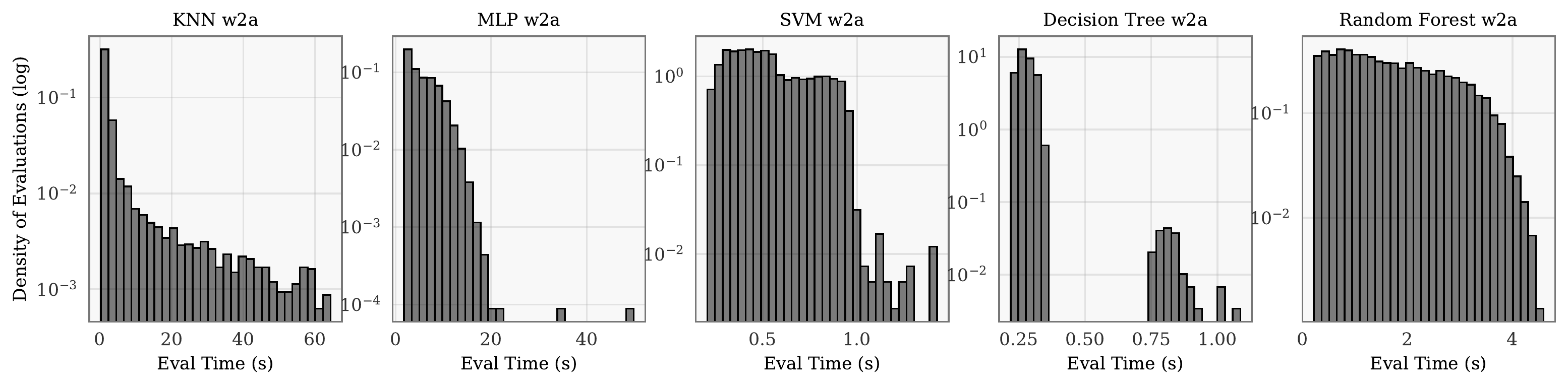}
    \caption{Runtime distribution, log-scaled, of 5000 randomly selected points for the K-nearest-neighbors (KNN), Multi-layer Perceptron (MLP), Support Vector Machine (SVM), Decision Tree (DT), and Random Forest (RF) hyperparameter optimization problems, each trained on the w2a dataset. The runtimes vary, often by an order of magnitude or more.}
    \label{fig:variations}
\end{figure*}

BO's sample efficiency leads to fast convergence only if evaluations cost the same, an assumption that is often not true in practice. Figure~\ref{fig:variations} illustrates this by randomly evaluating 5000 hyperparameter configurations for five popular HPO problems. Unsurprisingly, resulting evaluation times vary, often by an order of magnitude or more. Moreover, the bulk of each problem's search space tends to be cheap, suggesting significant cost savings may be achieved by using a cost efficient rather than a sample efficient optimizer. 

Motivated by this, we aim to make BO \textit{cost-aware}. This cost may be time, energy, or money, and the goal is to minimize the objective given a cost budget. We first illustrate the novel challenges behind cost-aware BO, and explain why current methods are not adequate. Then, we introduce \textbf{C}ost \textbf{A}ppo\textbf{r}tioned \textbf{BO} (CArBO), a novel BO algorithm that combines a cost-effective initial design with a budget-aware acquisition function, which can be run both sequentially and in batch. We show in an extensive set of experiments drawn from 20 real-world HPO problems that CArBO significantly outperforms competing methods within the same budget. Finally, we design low-variance cost models that extrapolate well, and demonstrate that they can further improve the performance of cost-aware BO.

\section{Background and Related Work}\label{Related Work}
Most prior approaches to cost-aware BO occur in the \textit{grey-box} setting, in which additional information about the objective is available. \textit{Multi-fidelity} BO is one such widely studied approach in which fidelity parameters $s \in [0, 1]^m$, such as iteration count or grid size, are assumed to be a noisy proxy for high-fidelity evaluations \cite{forrester2007multi, kandasamy2017multi, poloczek2017multi, wu2019practical}. Increasing $s$ decreases noise at the expense of runtime. Multi-fidelity methods are often application-specific. For example, Hyperband \cite{li2017hyperband} and its BO variants \cite{falkner2018bohb, klein2016fast, klein2017fastbo} cheaply train many neural network configurations for only a few epochs, and then train a selected subset for further epochs. In \textit{multi-task} BO, hyperparameter optimization is run on cheaper training sets before more expensive ones. \citet{swersky2013multi} introduce a cost-aware, multi-task variant of entropy search to speed-up optimization of logistic regression and latent Dirichlet allocation. Cost information is input as a set of cost preferences (e.g., parameter $\xb_1$ is more expensive than parameter $\xb_2$) by  \citet{abdolshah2019costaware}, who develop a multi-objective, constrained BO method that evaluates cheap points before expensive ones, as determined by the cost preferences, to find feasible, low-cost solutions.

These prior methods outperform their black-box counterparts by evaluating cheap proxies or cheap points before carefully selecting expensive evaluations. This \textit{early and cheap, late and expensive} strategy is accomplished by leveraging additional cost information inside the optimization routine. While these methods demonstrate strong performance, they sacrifice generality and do not apply to black-box BO. To our knowledge, cost-aware BO in the general black-box setting has not been thoroughly investigated. The de-facto standard in this setting is to normalize the acquisition by a GP cost model \cite{snoek2012practical}. This extends EI to \textit{EI per unit cost (EIpu)}:
\begin{equation}
    \EIpu(\xb ) \coloneqq \frac{\EI(\xb)}{c(\xb )},
\end{equation}
which is designed to balance the objective's cost and evaluation quality. \citet{snoek2012practical} showed that EIpu can boost performance on a variety of HPO problems.

\begin{figure}[t]
    \centering
    \includegraphics[scale=0.5]{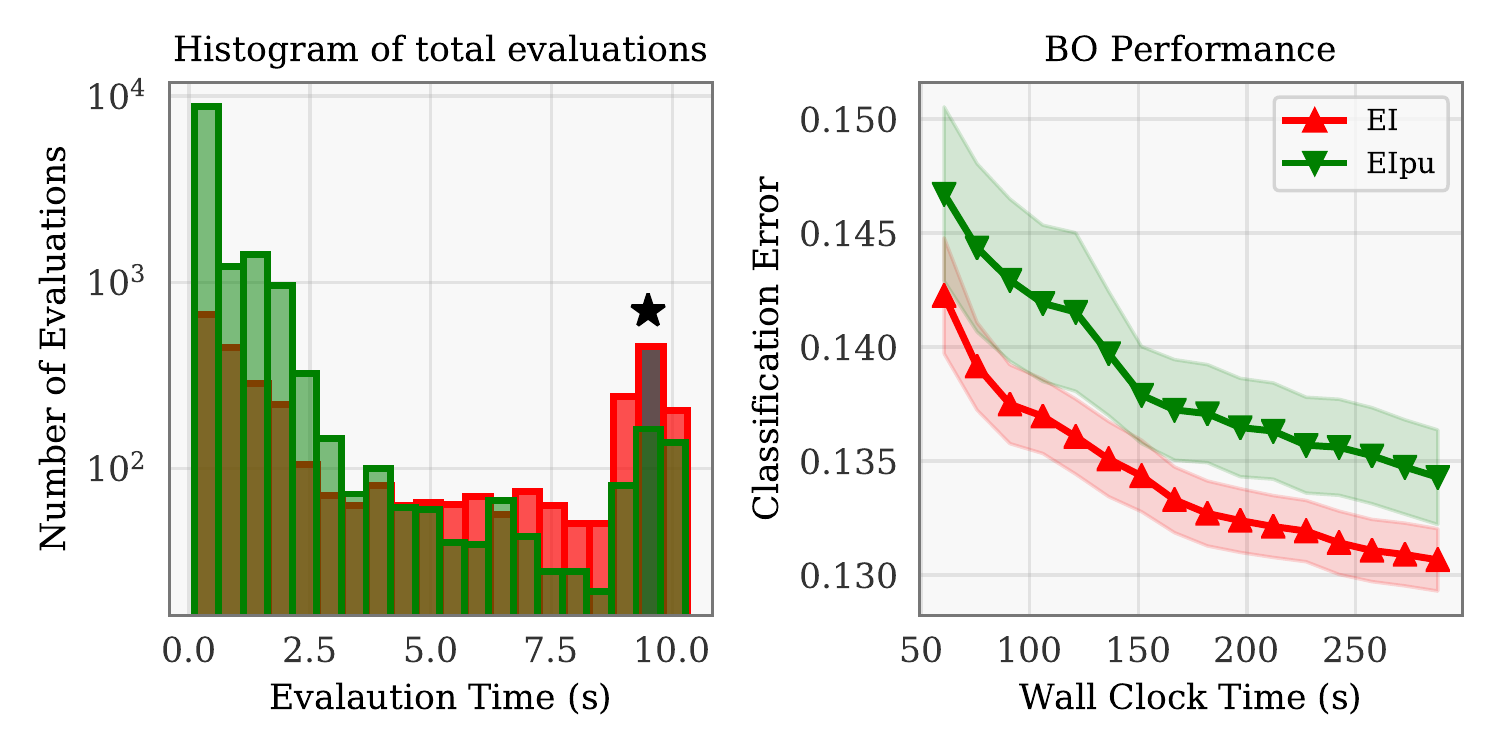}
    \caption{We run EI and EIpu on KNN, 51 times each. \textit{Left:} EIpu evaluates many more cheap points than EI, which evaluates more expensive points. The optimum's cost, one of the most expensive points, is a black star. \textit{Right:} EIpu performs poorly as a result. }
    \label{fig:knn_histogram}
\end{figure}

\begin{figure*}[t!]
    \centering
    \includegraphics[scale=0.6]{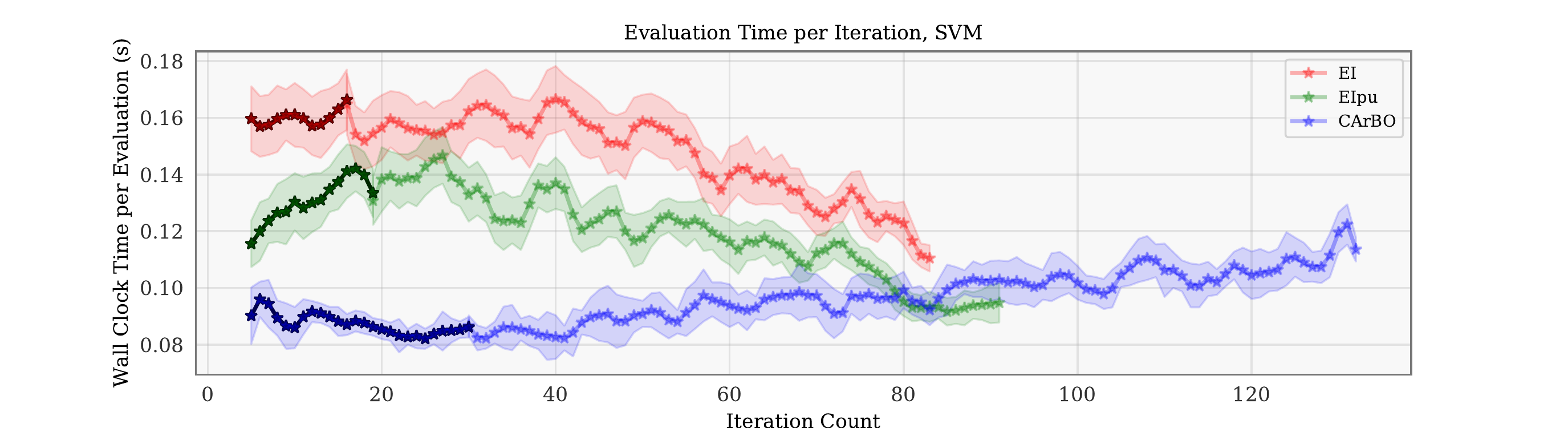}
    \caption{We plot the median evaluation time per iteration using each method's median number of iterations for an SVM HPO problem. We shade the iterations that consume the first $\tau/8$ cost, corresponding to the budget consumed by CArBO's initial design. CArBO clearly starts with many cheap evaluations and gradually evaluates more expensive points, enabling it to outperform EI and EIpu (see Section~\ref{experiments}).}
    \label{fig:optimization_timings}
\end{figure*}

In our benchmarks, EIpu demonstrated underwhelming performance. As we will show in Section~\ref{experiments}, out of twenty HPO problems, EIpu was worse than EI on nine. We illustrate EIpu's poor performance on certain problems in Figure \ref{fig:knn_histogram}, in which EIpu (green) is slower than EI (red) at HPO of a K-nearest-neighbor model. The empirical optimum, namely the best point over all trials (black star), has high cost. As a result, dividing by the cost penalizes EIpu \textit{away} from the optimum and diminishes its performance. This is evidenced by the evaluation time histograms: EIpu evaluates far more very cheap points compared to EI, which instead evaluates fewer but more expensive points. Due to its bias towards cheap points, EIpu and, more generally, dividing acquisition by cost, is likely to only display strong results when optima are relatively cheap. This is a problem in the black-box setting as we do not know the global optima's cost a priori. Intuitively, one can adversarially increase the optimum's cost to make EIpu underperform. We argue that a better cost-aware strategy can be introduced.

Standard BO is sequential but can be extended to the batch setting. In this context, $b$ candidates $\xb^1, \dots, \xb^b$ are selected by a batch acquisition function and then evaluated in parallel. Batch BO is inherently less sample efficient than sequential BO, and a batch acquisition seeks a set of diverse candidates that preserve sample efficiency. The challenge in batch BO is achieving linear scaling in $b$. Linear scaling means batch BO converges to the same optima as sequential BO with only $1/b$ iterations. Batch fantasizing extends any acquisition function to the batch setting by predicting multiple future evaluation trajectories, known as ``fantasies''\citep{wilson2018maximizing}. These fantasies are aggregated to identify a sequence of evaluation points, which is then evaluated in parallel.

For the rest of this paper, we measure cost with time. In the batch setting with fixed $b$, time is equivalent to both energy and money, as we assume each parallel resource consumes the same power or dollars per unit time. If batch size varies this is no longer true, but we leave this as future work. 

\section{CArBO: Cost Apportioned BO}\label{CArBO}

We introduce CArBO, an EI-based method employing the \textit{early and cheap, late and expensive} strategy from the multi-fidelity and multi-task setting. This strategy is seen in Figure~\ref{fig:optimization_timings}: in contrast with EI and EIpu, CArBO evaluates cheaper points before expensive ones. CArBO does this through a cost-effective initial design and cost-cooling. First, the cost-effective initial design aims to maximize coverage of the search space with cheap evaluations, building a good surrogate within a warm-start budget. Then, cost-cooling starts the optimization with EIpu and ends it with EI by deprecating the cost model as iterations proceed. We discuss each of these two building blocks next.

\textbf{Cost-effective initial design.} BO is always warm-started with an \textit{initial design}. A design is a set of points selected to learn variation in data, and BO evaluates an initial design before optimization starts to provide starting data for its GP. Initial designs consume some budget, and therefore must balance information gain with sample efficiency. An overly small design yields a poor surrogate, while an overly large design decreases sample efficiency. Initial designs must therefore be evenly spaced throughout the domain, and are often Latin hypercubes or low-discrepancy sequences \cite{kirk2012experimental, ryan2007modern}. These scale better than grid points and distribute more evenly than random points \cite{stein1987large, niederreiter1988low}. In practice, BO initial designs are small; the popular GPyOpt software uses five points \cite{gpyopt2016}, though seminal work suggests $10d$ points, where $d$ is the problem dimension \cite{jones1998ego}. 

In cost-aware BO we aim to design a \textit{cost-effective} initial design, which balances information gain with cost efficiency. A cost-effective design fills $\Omega$ with more evaluations than a traditional initial design within the same warm-start budget $\tau_{init}$. We select a cost-effective initial design through the following optimization subproblem:
\begin{equation}\label{eq:fill}
\begin{aligned}
& \argmin_{\Xb \in 2^{ \Omega}} 
& & \text{fill}(\Xb) \coloneqq  \sup_{\xb \in \Omega} \min_{\xb_j \in \Xb} \| \xb_j - \xb \|_2. \\
& \text{subject to}
& & \sum_{\xb_i \in \Xb} c(\xb_i) < \tau_{init}.
\end{aligned}
\end{equation}

Here, fill($\Xb$) is the radius of the largest empty sphere one can fit in $\Omega$. It measures the spacing of $\Xb$ in $\Omega$, and is known as the minimax criterion in the design-of-experiments literature \cite{pronzato2012design}. The smaller a set's fill is, the better distributed it is within $\Omega$. The argmin of Eq.\ (\ref{eq:fill}) is the initial design within $\tau_{init}$ cost with the smallest fill. 

\begin{algorithm}[h]
   \caption{Cost-effective initial design}
   \label{alg:initial_design}
\begin{algorithmic}[1]
   \STATE \textbf{Input}: initial budget $\tau_{init}$, optimization domain $\Omega$.
   \STATE Cumulative time $ct = 0$, initial design $\Xb_{init} = \{\}$.
    \STATE Discretize $\Omega$ into $\tilde \Omega$.
    \WHILE{$ct < \tau_{init}$ }
        \WHILE{size $\Xb_{cand} > 1$}
        \STATE exclude most expensive point from $\tilde \Omega$.
        \STATE exclude point closest to $\Xb_{init}$ from $\tilde \Omega$.
        \ENDWHILE
        \STATE add remaining point to $\Xb_{init}$ and evaluate.
        \STATE Update $ct$, cost surrogate. 
    \ENDWHILE
    \STATE \textbf{return} $\Xb_{init}$.
\end{algorithmic}
\end{algorithm}

\begin{figure}[t!]
    \centering
    \includegraphics[scale=0.5]{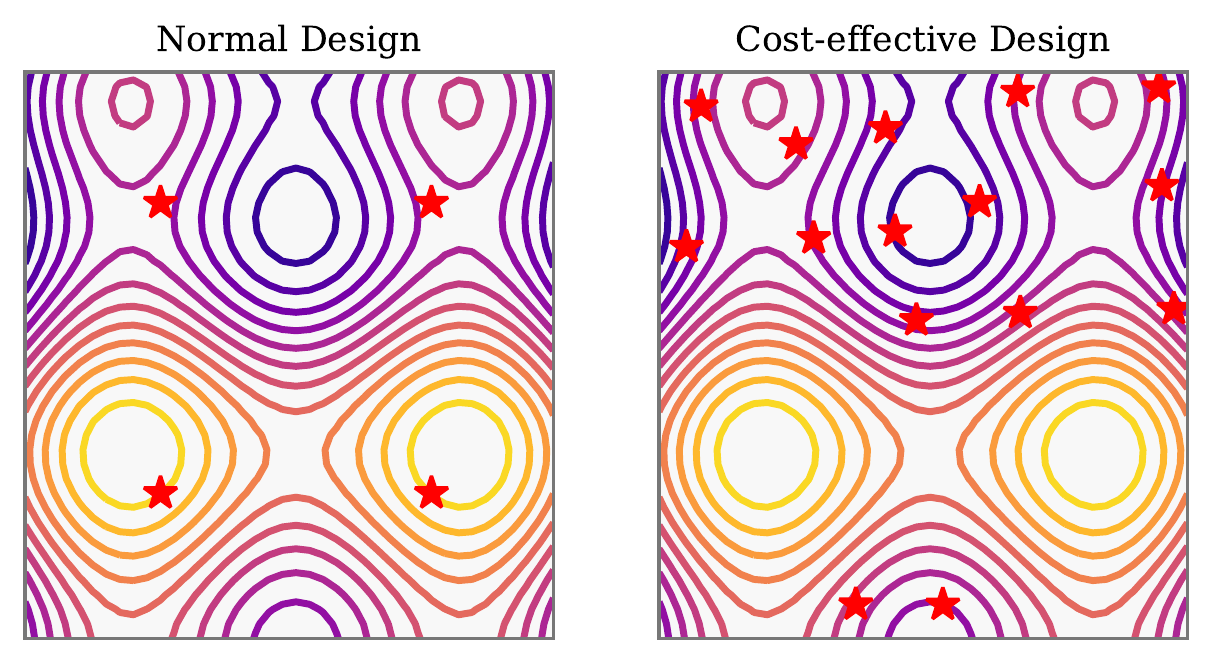}
    \caption{Two initial designs with the same cost, plotted over a contour of the synthetic cost function. \textit{Left: }a grid of four points. \textit{Right: }a cost-effective solution containing 15 points, which covers the search space better than the grid.}
    \label{fig:initialdesign}
\end{figure}

Eq.\ (\ref{eq:fill}) is a difficult optimization problem. In the discrete setting with constant cost, it an instance of the vertex cover problem, known to be NP-complete. Typically, approximations to the minimax criterion are built greedily, and have a worst-case approximation factor of 2 \cite{damblin2013numerical, pronzato2017minimax}. Algorithm \ref{alg:initial_design} is a variation of these approaches for non-constant cost functions, and reduces to the greedy approach described in \citet{pronzato2012design} given a constant cost function. Algorithm \ref{alg:initial_design} first discretizes $\Omega$ into candidates $\tilde \Omega$ and then adds a point from $\tilde \Omega$ to the initial design as follows: remove the highest cost point and then the shortest distance point from  $\tilde \Omega$, continuing until only one point remains. This remaining candidate is cheap and far from other points in the design. This inner loop is repeated, updating $c(\xb)$ every iteration until $\tau_{init}$ is exceeded. This results in a set of cheap and well-distributed points. In the batch setting, the inner loop is run $b$ times to select $b$ candidates that are then evaluated in parallel. Figure~\ref{fig:initialdesign} shows that a cost-effective design gains far more information than a standard grid, with fifteen points compared to four. 

\textbf{Cost-cooling.} The second building block is cost-cooling. Assume that at the $k$th BO iteration, $\tau_k$ of the total budget $\tau$ has been used (at $k=0$, $\tau_k = \tau_{init}$). Cost-cooling, which we call EI-cool when using EI, is defined as:
\begin{equation}\label{ei-cool}
\text{EI-cool}(\xb) \coloneqq \frac{\EI(\xb)}{c(\xb)^\alpha} \;,\; \alpha = (\tau - \tau_k) / (\tau - \tau_{init}).
\end{equation}
Cost $c(\xb)$ is assumed to be positive and modeled with a warped GP that fits the log cost $\gamma(\xb)$. The cost is predicted by $c(\xb) = \exp (\gamma(\xb))$ as in the standard EIpu \cite{snelson2004warped, snoek2012practical}. We discuss alternative low-variance cost models that extrapolate well in Section \ref{Cost Models}. Learning $c(\xb)$ requires a warm-start, for which we use five points drawn from the search space uniformly at random. 

As the parameter $\alpha$ decays from one to zero, EI-cool transitions from EIpu to EI. As a result, cost-cooling de-emphasizes the cost model as the optimization progresses and cheap evaluations are performed before expensive ones. As mentioned earlier, this behavior is shown by Figure~\ref{fig:optimization_timings}, and also by additional benchmarks located in the appendix. The idea of cost-cooling bears connections to previous work on multi-objective, cost-preference BO \cite{abdolshah2019costaware}, where cost constraints are loosened to ensure that the entire Pareto frontier is explored. As we show in the appendix, EI-cool is not guaranteed to outperform both EIpu and EI but usually outperforms \textit{at least one}. 

\begin{figure*}[ht]
    \centering
    \includegraphics[width=\textwidth]{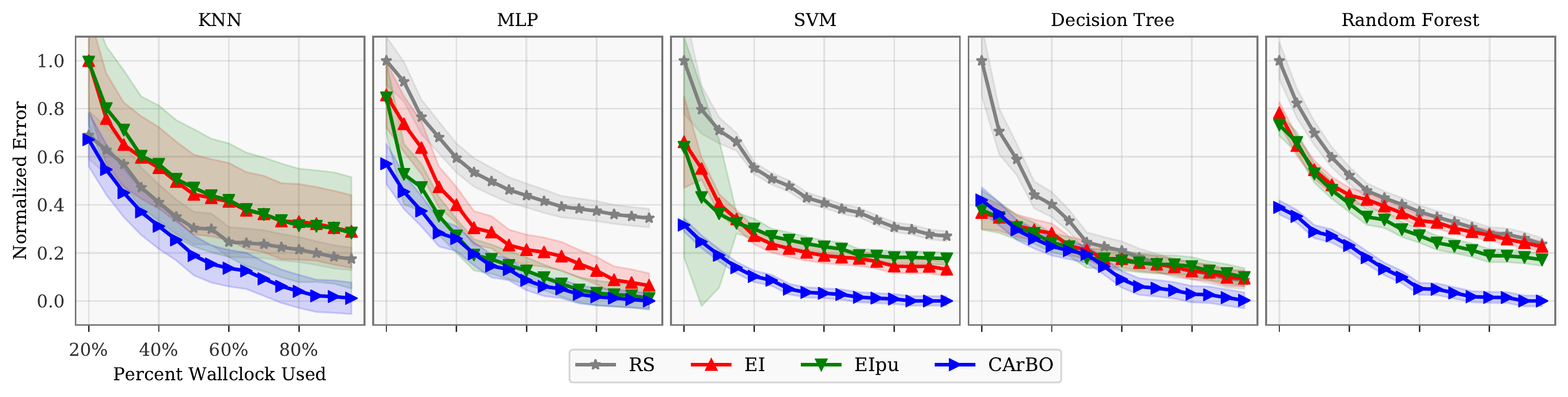}
    \includegraphics[width=\textwidth]{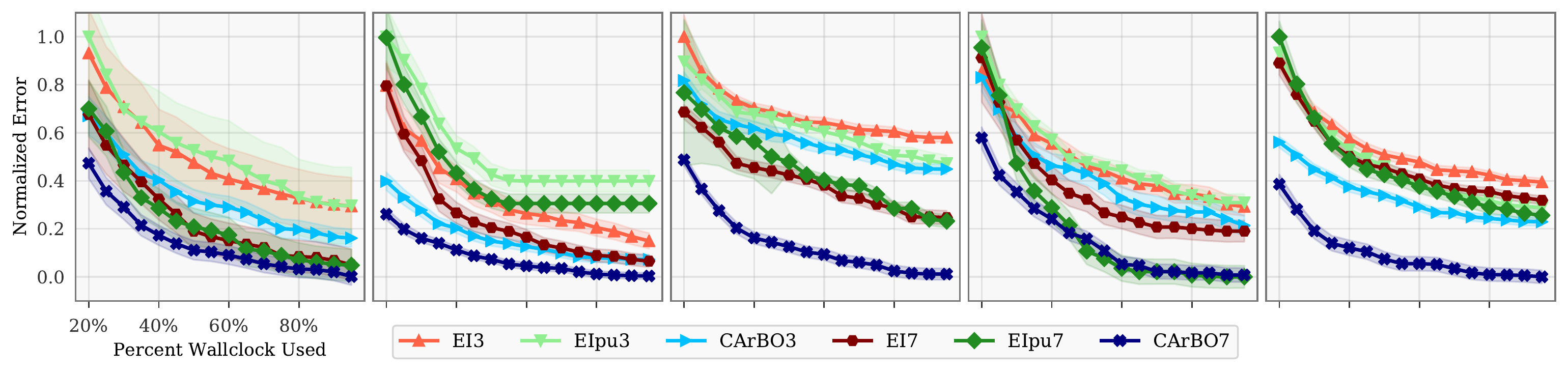}
    \caption{\textit{Top:} Sequential comparison. \textit{Bottom:} Batch comparison, of batch sizes 3 and 7. RS is plotted in grey, EI methods are plotted in red, EIpu methods are plotted in green, and CArBO methods are in plotted blue. In almost all cases, CArBO converges significantly faster than competing methods. The median is plotted, with one standard deviation shaded above and below. }
    \label{fig:sklearn_results}
\end{figure*}

\textbf{CArBO.} The overall method we propose is detailed in Algorithm \ref{alg:CArBO}, which combines the cost-effective initial design and cost-cooling. 
For its default cost-effective design budget we use $\tau_{init} = \tau / 8$, where $\tau$ denotes the total BO budget. We found this value of $\tau_{init}$ to work well in experiments, and investigate different choices in Section~\ref{Ablation study}. 

We formulate CArBO in the general batch setting, namely with $b \geq 1$. Performing evaluations in parallel further reduces wall clock time, and can be achieved with standard techniques as described with pseudo-code in the appendix. Note that in the cost-aware setting, linear scaling means convergence to the same optimum as sequential BO in $1/b$ of the wall clock time. CArBO achieves linear scaling  by building its cost-efficient initial design with batches of points that are far apart from each other and then by batch-fantasizing cost-cooling. We will demonstrate this scaling on relatively large batch sizes of up to 16 in Section~\ref{Ablation study}. 

\begin{algorithm}[t!]
   \caption{CArBO: Cost Apportioned BO}
   \label{alg:CArBO}
\begin{algorithmic}[1]
   \STATE \textbf{Input}: batch $b$,  initial budget $\tau_{init}$, budget $\tau$
   \STATE Cumulative time $ct = 0$.
   \STATE Evaluate cost-effective initial design via Algorithm~\ref{alg:initial_design} using $\tau_{init}$.
    \STATE Update $ct$, cost and objective surrogates.
    \WHILE{ $ ct < \tau$}
        \STATE $\xb^1, \dots, \xb^b \gets $ EI-cooling batch fantasy $b$ as per Eq.~\ref{ei-cool}.
        \STATE Evaluate $\xb^1, \dots, \xb^b$ in parallel.
        \STATE Update $ct$, objective and cost surrogates. 
    \ENDWHILE
    \STATE \textbf{return} best hyperparameter configuration observed.
\end{algorithmic}
\end{algorithm}

\section{Empirical Evaluation}\label{experiments}

\begin{table*}[ht]
\centering 
\scriptsize
\setlength\tabcolsep{4pt}
\begin{tabular}{ | l  l  l l l l l l l l l | }
\hline 
Objective & Budget & EI3 & EI7 & EI11 & EIpu3 & EIpu7 & EIpu11 & CArBO3 & CArBO7 & CArBO11 \\
\hline \hline
KNN a1a 	& 150 (s) &    0.133 (83) & 0.128 (149) & \textbf{0.126} (238) & 0.135 (121) & 0.128 (195) & 0.126 (318) & 0.133 (111) & 0.128 (250) & 0.128 (411) \\\hline
KNN a3a 	& 300 (s) &    0.121 (90) & 0.117 (184) & 0.115 (283) & 0.121 (116) & 0.117 (217) & 0.115 (331) & 0.119 (147) & 0.116 (354) & \textbf{0.115} (622) \\\hline
KNN splice 	& 10 (s)  & 0.123 (143) & 0.107 (275) & 0.099 (411) & 0.120 (183) & 0.107 (361) & 0.102 (536) & 0.113 (161) & 0.103 (353) & \textbf{0.095} (537) \\\hline
KNN w2a 	& 400 (s) &    0.055 (83) & 0.052 (150) & 0.047 (206) & 0.056 (142) & 0.049 (277) & 0.048 (373) & 0.048 (77) & 0.046 (189) & \textbf{0.044} (314) \\\hline
MLP a1a 	& 100 (s)  &    0.123 (50) & 0.122 (96) & 0.122 (133) & 0.128 (34) & 0.127 (72) & 0.126 (103) & 0.121 (119) & 0.119 (227) & \textbf{0.119} (344) \\\hline
MLP a3a 	& 160 (s)  &    0.108 (40) & 0.108 (79) & 0.107 (114) & 0.110 (30) & 0.108 (62) & 0.108 (90) & 0.107 (97) & 0.106 (194) & \textbf{0.106} (296) \\\hline
MLP splice 	& 50 (s)  & 0.051 (41) & 0.043 (84) & 0.041 (126) & 0.054 (32) & 0.052 (64) & 0.050 (92) & 0.038 (71) & 0.037 (145) &\textbf{ 0.036} (215) \\\hline
MLP w2a 	& 200 (s) &    0.024 (33) & 0.023 (69) & 0.022 (101) & 0.024 (27) & 0.023 (57) & \textbf{0.022} (84) & 0.023 (73) & 0.023 (152) & 0.023 (226) \\\hline
SVM a1a 	& 20 (s)  &    0.120 (189) & 0.120 (395) & 0.120 (587) & 0.120 (218) & 0.120 (483) & 0.120 (753) & 0.120 (295) & 0.119 (663) & \textbf{0.119} (956) \\\hline
SVM a3a 	& 30 (s)  &    0.109 (197) & 0.108 (418) & 0.108 (611) & 0.108 (256) & 0.107 (572) & 0.107 (913) & 0.107 (343) & 0.107 (722) & \textbf{0.106} (1019) \\\hline
SVM splice 	& 4 (s)   & 0.114 (100) & 0.114 (191) & 0.113 (282) & 0.114 (127) & 0.113 (307) & 0.113 (425) & 0.113 (225) & 0.111 (540) & \textbf{0.111} (836) \\\hline
SVM w2a 	& 90 (s)  &    0.023 (256) & 0.022 (570) & 0.022 (855) & 0.022 (304) & 0.021 (676) & 0.021 (1040) & 0.021 (356) & 0.021 (763) & \textbf{0.020} (1034) \\\hline
DT a1a 		& 2.5 (s) &     0.135 (150) & 0.132 (347) & \textbf{0.132} (541) & 0.135 (149) & 0.132 (347) & 0.132 (537) & 0.135 (150) & 0.132 (344) & 0.132 (540) \\\hline
DT a3a 		& 2.5 (s) &     0.132 (133) & 0.130 (300) & 0.129 (473) & 0.132 (135) & 0.129 (300) & 0.130 (464) & 0.131 (134) & 0.130 (304) & \textbf{0.128} (476) \\\hline
DT splice 	& 2 (s)   &  0.029 (300) & 0.028 (645) & 0.026 (1032) & 0.029 (300) & 0.025 (655) & 0.027 (979) & 0.029 (332) & 0.027 (664) &\textbf{ 0.025} (985) \\\hline
DT w2a 		& 8 (s)   &     0.055 (77) & 0.077 (177) & 0.078 (277) & 0.052 (80) & 0.078 (181) & 0.078 (279) & 0.054 (78) & 0.054 (173) & \textbf{0.052} (272) \\\hline
RF a1a 		& 30 (s)  &     0.117 (68) & 0.116 (137) & 0.116 (214) & 0.116 (133) & 0.115 (270) & 0.114 (373) & 0.116 (160) & 0.114 (272) & \textbf{0.114} (359) \\\hline
RF a3a 		& 35 (s)  &     0.110 (80) & 0.108 (170) & 0.108 (248) & 0.109 (118) & 0.109 (243) & \textbf{0.108} (337) & 0.109 (143) & 0.108 (252) & 0.108 (355) \\\hline
RF splice 	& 10 (s)  &  0.015 (31) & 0.013 (73) & 0.013 (110) & 0.015 (55) & 0.013 (114) & 0.013 (162) & 0.014 (46) & 0.013 (88) & \textbf{0.012} (118) \\\hline
RF w2a 		& 80 (s)  &     0.049 (60) & 0.053 (258) & 0.051 (389) & 0.045 (135) & 0.053 (312) & 0.051 (484) & 0.044 (142) & 0.042 (298) & \textbf{0.041} (383) \\\hline
\end{tabular}
\caption{Results for all different batch methods on five HPO tasks, each tested on four datasets using 51 replications. The tasks are K-nearest-neighbors (KNN), multi-layer perceptron (MLP), support-vector machine (SVM), decision tree (DT), and random forest (RF). The datasets are a1a, a3a, splice, w2a. The median classification error is shown for different optimizers and batch sizes. CArBO11 displays strong results, showing the best on 16 out of the 20 benchmarks and lagging by a small amount in the other four cases. } \label{tab:batch_results}
\end{table*}

We evaluate the performance of CArBO on a varied set of five popular HPO problems, each trained on four different datasets, yielding twenty total benchmarks. Each benchmark is given its own wall clock budget. Each HPO problem is a model in scikit-learn \cite{scikit-learn}. We train on four classification datasets: \textit{splice}, \textit{a1a}, \textit{a3a}, and \textit{w2a}. The splice dataset (training size: 1000, testing size: 2175) classifies splice junctions in a DNA sequence. The a1a and a3a datasets (training sizes: 1605, 2265, testing sizes: 30,956, 30,296) predict whether the annual income of a family exceeds 50,000 dollars based on 1994 US census data. The w2a dataset (training size: 3,470, testing size: 46,279) predicts the category of a webpage. All datasets are available in the UCI machine learning repository \cite{Dua:2019}. Each benchmark is replicated 51 times on independent AWS m4.xlarge machines to ensure consistent evaluation times. The problems and search spaces follow, with unlisted hyperparameters being set to the scikit-learn default. 

\textbf{K-nearest-neighbors (KNN).} We consider a 5$d$ search space: dimensionality reduction percentage and type in [1e-6, 1.0] log-scaled and \{Gaussian, Random\}, respectively, neighbor count in \{1, 2, ..., 256\}, weight function in \{Uniform, Distance\}, and distance in \{Minkowski, Cityblock, Cosine, Euclidean, L1, L2, Manhattan\}. 

\textbf{Multi-layer perceptron (MLP).} We consider a 11$d$ search space: number of layers in \{1, 2, 3, 4\}, layer sizes in \{10, 11, ..., 150 \} log-scaled, activation in \{Logistic, Tanh, ReLU\}, tolerance in [1e-5, 1e-2] log-scaled, and Adam parameters \cite{kingma2014adam}: step size in [1e-6, 1.0] log-scaled, initial step size in [1e-6, 1e-2] log-scaled, beta1 and beta2 in [1e-3, 0.99] log-scaled.

\textbf{Support Vector Machine (SVM).} We consider a 6$d$ search space: iteration count in \{1, 2, ..., 128\}, penalty term in \{L1, L2, ElasticNet\}, penalty ratio in [0, 1], step size in [1e-3, 1e3] log-scaled, initial step size in [1e-4, 1e-1] log-scaled, optimizer in \{Constant, Optimal, Invscaling, Adaptive \}. 

\textbf{Decision tree (DT).} We consider a 3$d$ search space: tree depth in \{1, 2, ..., 64\}, tree split threshold in [0.1, 1.0] log-scaled, and split feature size in [1e-3, 0.5] log-scaled. 

\textbf{Random forest (RF).} We consider a 3$d$ search space: number of trees in \{1, 2, ..., 256\}, tree depth in \{1, 2, ..., 64\}, and tree split threshold in [0.1, 1.0] log-scaled. 

Our code is built on GPyOpt \cite{gpyopt2016}. Kernel hyperparameters for both the objective and cost Gaussian process models are calculated via maximum marginal likelihood estimation \cite{rasmussen2006gaussian}. We compare CArBO to EI, EIpu, and random search in the sequential case, as well as batch sizes three, seven, and eleven. Note that multi-fidelity methods such as Hyperband are inapplicable on these benchmarks as they do not have any fidelity parameters (with the exception of SVM). 

\begin{table}[t]
\centering
\scriptsize
\setlength\tabcolsep{4pt}
\begin{tabular}{| l  c c c c |}
\hline
Objective & CArBO & CArBO3 & CArBO7 & CArBO11 \\
\hline \hline
KNN a1a &     60\% &  49\% & -10\% & -11\% \\ \hline		
KNN a3a &     52\% &  58\% &  22\% &  28\% \\ \hline
KNN splice &  73\% &  75\% &  52\% &  49\% \\ \hline
KNN w2a &     59\% &  55\% &  60\% &  59\% \\ \hline
MLP a1a &     21\% &  69\% &  67\% &  69\% \\ \hline
MLP a3a &     -9\% &  50\% &  61\% &  56\% \\ \hline
MLP splice &  34\% &  62\% &  66\% &  59\% \\ \hline
MLP w2a &     4\%  &  27\% &  20\% &  -7\% \\ \hline
SVM a1a &     22\% &  42\% &  53\% &  39\% \\ \hline
SVM a3a &     67\% &  66\% &  65\% &  52\% \\ \hline
SVM splice &  -1\% &  50\% &  67\% &  67\% \\ \hline
SVM w2a &     74\% &  78\% &  22\% &  72\% \\ \hline
DT a1a &      -2\% &  -7\% &  17\% &  -8\% \\ \hline
DT a3a &      15\% &  22\% & -22\% &  35\% \\ \hline
DT splice &   10\% &   2\% & -25\% &   2\% \\ \hline
DT w2a & -    18\% & -41\% &  95\% &  96\% \\ \hline
RF a1a &      44\% &  28\% &  63\% &  61\% \\ \hline
RF a3a &      40\% &  54\% &  49\% & -24\% \\ \hline
RF splice &   16\% &  33\% &  27\% &  33\% \\ \hline
RF w2a &      52\% &  48\% &  82\% &  84\% \\ \hline
\textbf{Net Saving} &      \textbf{32.5}\% &  \textbf{45.1}\% &  \textbf{41.6}\% &  \textbf{40.6}\% \\
\hline
\end{tabular}
\caption{For each batch size and objective, we calculate the median cost savings as a percentage of budget. Negative numbers indicate that CArBO performed worse than the best optimizer. CArBO performs strongly on the large majority of problems. Furthermore, when it does worse, it only does worse by a small amount.} \label{tab:percentage_savings}
\end{table}

We compare the performance of the competing algorithms in three ways. First, we plot performance for each HPO problem by averaging the classification error for each model over the four datasets used (Figure \ref{fig:sklearn_results}). This is done to condense the large number of benchmarks we ran. We average as follows: first we normalize performance so that the worst optimizer starts optimization at 1.0 and the best optimizer ends at 0.0, then we take the mean over all datasets. We plot sequential results in the first row and batch results in the second. CArBO outperforms both EI and EIpu by a large margin across all batch sizes.

Second, we compile a table of classification errors and iterations taken, and bold the optimizer with the lowest classification error (Table~\ref{tab:batch_results}). For space's sake, we truncate the classification error precision to three digits. The table also lists each benchmark's time budget in seconds. CArBO for batch sizes one, three, seven, and eleven is best on 16, 18, 17, and 16 HPO problems, respectively. As expected, CArBO is able to exploit more BO iterations than either EI or EIpu for the same wall-clock time. 

Third, we calculate CArBO's total cost savings, defined as the time needed by CArBO to achieve comparable results to the next best optimizer (Table~\ref{tab:percentage_savings}). We consider Table~\ref{tab:percentage_savings} the most instructive comparison because it provides quantitative savings instead of a qualitative ranking. We list the median cost savings for each benchmark, as well as net savings over all benchmarks, for each batch size. CArBO achieves large cost savings of roughly 40 percent, averaged over all benchmarks and batch sizes.

\section{Additional Experiments} \label{Ablation study}
This section illustrates the empirical behavior of CArBO relative to its internal design choices, such as batch size or initial design budget. First, we investigate the sensitivity of CArBO to its initial design budget. We also run a scaling test for batch sizes up to 16. Finally, we run an ablation study. All following experiments use the MLP a1a benchmark. 

\begin{figure}[t]
    \centering
    \includegraphics[scale=0.5]{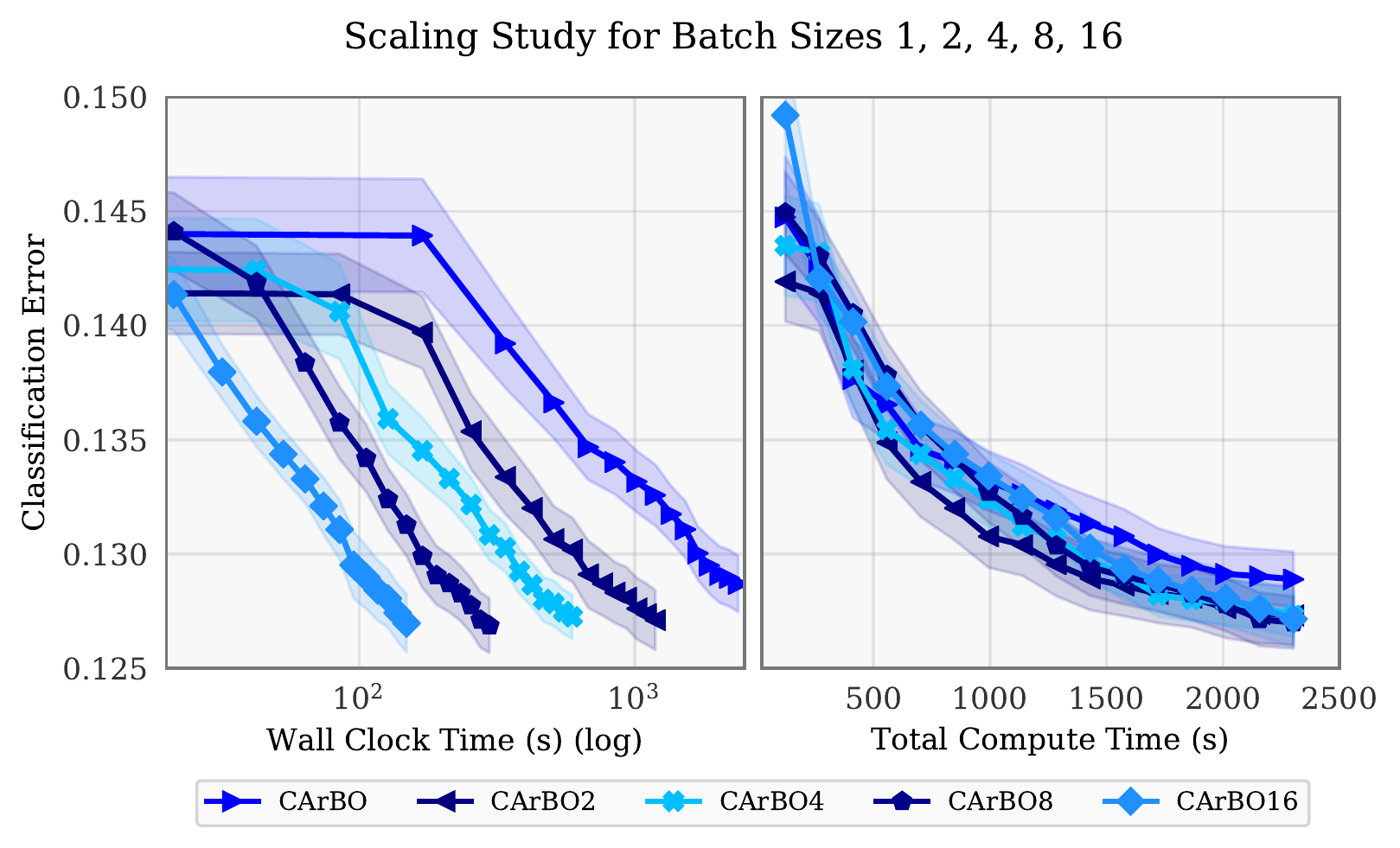}
    \caption{We compare CArBO's wall clock time performance (\textit{left}) to its total compute time performance (\textit{right}) for batch sizes 1, 2, 4, 8, and 16. CArBO scales linearly with batch, evidenced by comparable total compute time performance among all batch sizes.}
    \label{fig:scaling_study}
\end{figure}
\textbf{Batch Scaling. }Information is used less optimally in batch BO than in sequential BO. Large batches size may result in decreased cost efficiency. We examine this potential risk by running CArBO for batch size 1, 2, 4, 8, and 16 with wall clock time budgets of 2400, 1200, 600, 300, and 150 seconds, respectively. Each batch thus is allocated the same total compute time. As seen in Figure~\ref{fig:scaling_study}, moving up to batch size 16 results in little to no performance loss, indicating that CArBO scales linearly with batch size. 

\begin{figure}[h]
    \centering
    \includegraphics[scale=0.6]{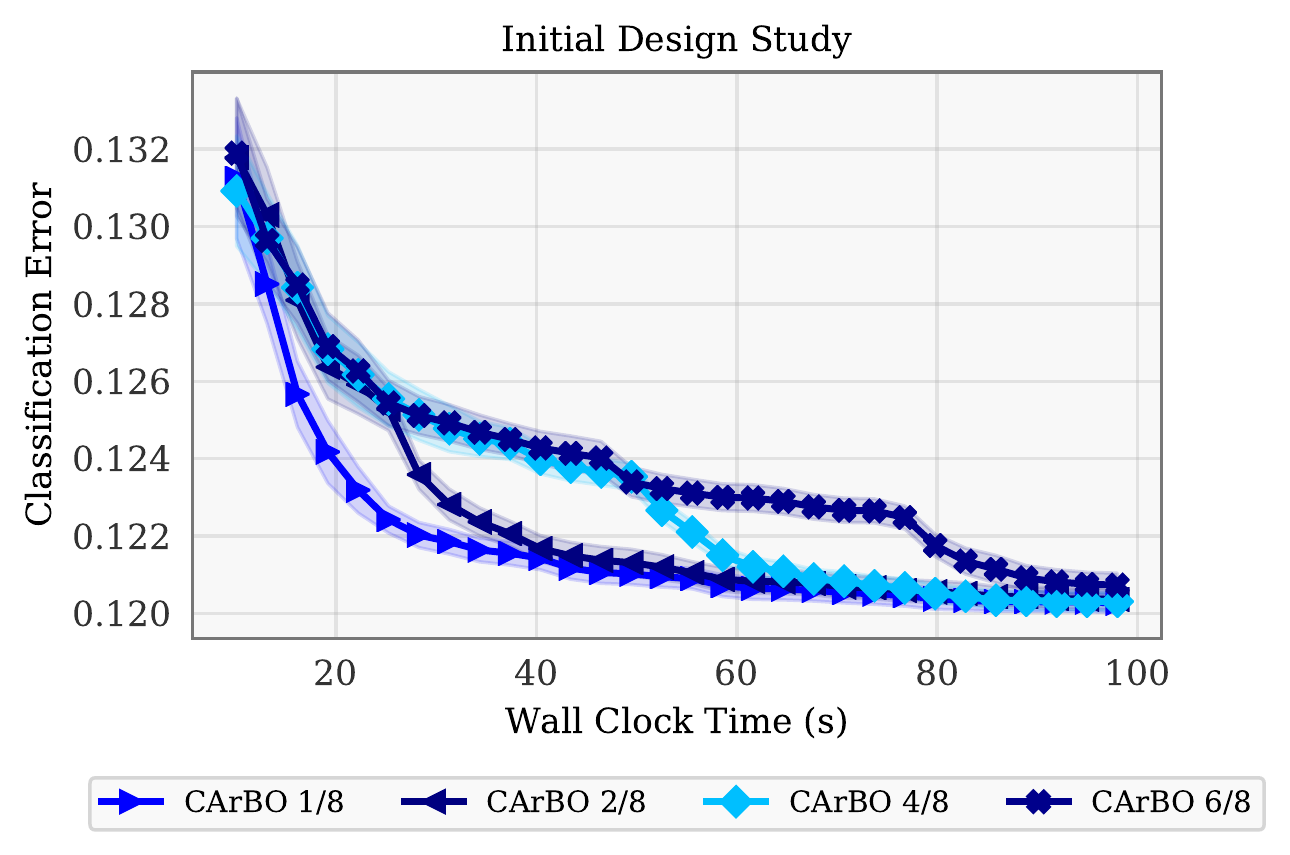}
    \caption{We study CArBO's initial design budget from 1/8 to 6/8 of the total budget. While CArBO 6/8 does perform worse, there is relatively little performance change, indicating at least some robustness to the initial design budget.}
    \label{fig:ratio_study}
\end{figure}
\textbf{Initial Design.} In BO methods, the size of the initial design is somewhat arbitrary. This is also true for CArBO, which uses 1/8 of the budget for the initial design. We investigate the impact of varying the initial design budget in Figure~\ref{fig:ratio_study} from 1/8 up to an extreme value of 6/8 of the total budget. CArBO's performance was relatively unchanged; using 6/8 of the total budget for the initial design degraded performance slightly, but represents an extreme case. We leave a systematic approach to select the initial design budget as future work.

\begin{figure}[h]
    \centering
    \includegraphics[scale=0.6]{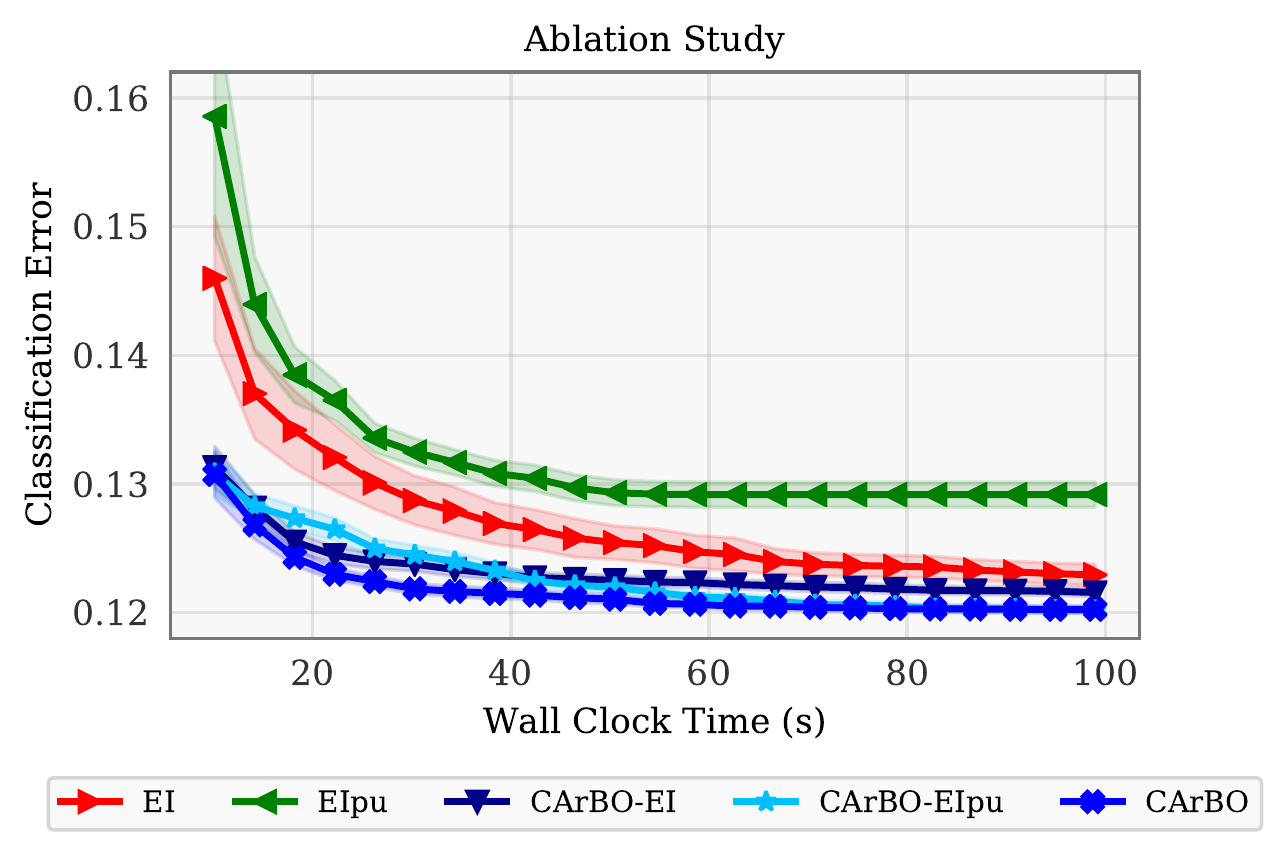}
    \caption{The cost-effective design contributes the larger performance increase compared to EI-cooling in this ablation study.}
    \label{fig:ablation_study}
\end{figure}
\textbf{Ablation Study. }CArBO combines two components: a cost-effective design and cost-cooling. A natural question is the performance contribution of each. To answer this question we perform an ablation study, in which we remove each component and re-run optimization. In Figure~\ref{fig:ablation_study}, we compare CArBO to CArBO using just EI or EIpu. We also compare these to EI and EIpu. The initial design contributes the larger performance increase, which is not surprising. At the same time, CArBO using EI-cooling performs the best.

\section{Cost Surrogates}\label{Cost Models}
Predicting the cost of a computer program is well-studied by prior work \cite{huang2010predicting, hutter2014algorithm, di2013adaptive, yang2018oboe, priya2011predicting}, in which a cost model forecasts system loads, dispatches computational resources, or determines computational feasibility. Gaussian processes extrapolate poorly, leading to high-variance cost predictions far away from data. This high-variance may introduce extra error that decreases BO performance. 

We show that cost-aware BO can benefit from specialized, low-variance cost models that extrapolate well. Floating point operations (flops) are a standard measure of a computer program's cost \cite{peise2012performance}. We consider a linear model that uses a small feature set, where each feature counts the flops of a subroutine in the program. The total runtime is modeled as a linear combination of these features. We train the model through robust regression with the Huber loss to deal with outliers in timing data \cite{boyd2004convex}. 
We model MLPs and convolutional neural network (CNNs) with low-variance cost models and show that they tend to improve BO performance.

\textbf{Multi-layer Perceptron.} Consider an MLP with layer sizes $n_1, n_2, \dots, n_k$. We define the following features: the cost of all matrix multiplications $\phi_{quad} = (n_1 n_2 + n_2 n_3 + \dots + n_{k-1} n_k)$ and the cost of batch normalization and activation functions $\phi_{linear} = (n_1 + n_2 + \dots + n_k)$. Batch size $b$ and epoch $e$ are constants, and are omitted. 
Our cost model is:
$$c(x) = c_1 \phi_{quad} + c_2 \phi_{linear} + c_3.$$ 

We let $k = 4$, $10 \leq n_i \leq 300$, $b = 100$, $e=200$, and use no dropout. We train our low-variance model on timing data consistent with EIpu's evaluations, and run EIpu with this model on all four MLP benchmarks 51 times each. The low-variance cost model improves BO performance. Median classification error and cost savings are shown in Table~\ref{table:MLP}.

\begin{figure}[t]
    \centering
    \includegraphics[scale=0.5]{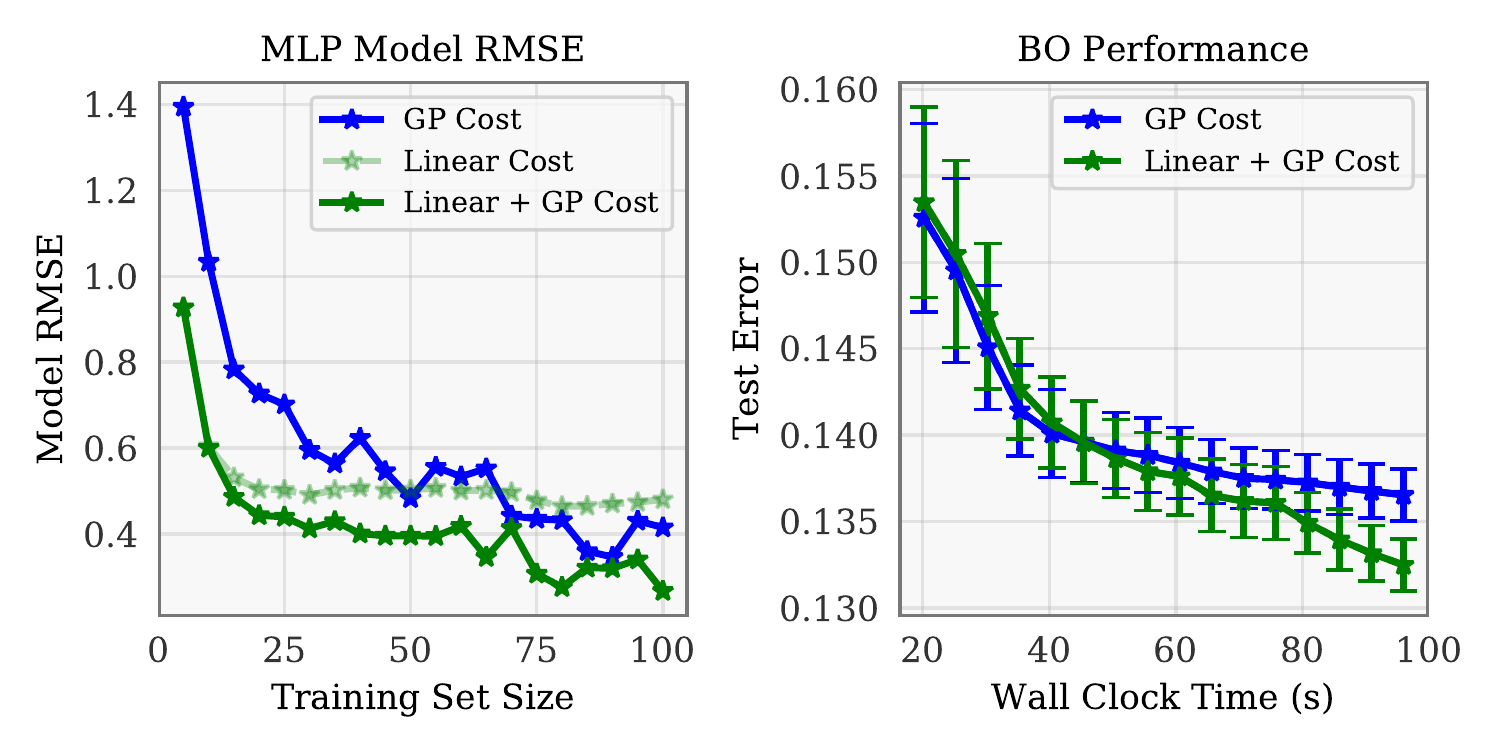}
    \caption{We run EIpu using both low-variance and warped GP models on MLP a1a. The warped GP (blue) has higher prediction error and slower performance than the low-variance model (green).}
    \label{fig:mlp_performance}
\end{figure}

\begin{table}[t]
\centering
\small
\setlength\tabcolsep{4pt}
\begin{tabular}{|l c c c | }
\hline
Objective  & Warped GP & Linear Model & Net Saving \\
\hline \hline
MLP a1a 	& 0.1363 & \textbf{0.1276} & 42\% \\ \hline
MLP a3a 	& 0.1124 & \textbf{0.1115} & 12\% \\ \hline
MLP splice  & 0.0686 & \textbf{0.0629} & 25\% \\ \hline
MLP w2a 	& 0.0257 & \textbf{0.0242} & 29\% \\ \hline
\textbf{Net Saving} & - & - & \textbf{27\%} \\ \hline
\end{tabular}
\caption{EIpu results with warped GP and low-variance models.}\label{tab:lowvariance}
\label{table:MLP}
\end{table}

The left plot of Figure~\ref{fig:mlp_performance} compares root-mean-squared error (RMSE) among three cost models: a warped GP (blue), our low-variance model (dotted green), and GP whose mean is our low-variance model (green), on 10,000 randomly chosen hyperparameters. A GP with a low-variance linear mean is the most accurate, while the warped GP is least accurate. Our low-variance models are strongest in the limited data regime; as the training set grows, the error gap shrinks. The right plot of Figure~\ref{fig:mlp_performance} shows significant improvement over the default cost model on the MLP a1a benchmark, which is the best performing benchmark out of the four we ran.

\textbf{Convolutional Neural Network.} Consider a CNN with $h$ convolutional layers of kernel size $r$, channel sizes $m_1, \dots, m_k$, pooling ratios $p_1, \dots, p_k$, a fixed activation function, and an input of size $I \times I \times c$, where $c$ is the number of color channels. We define the additional features: the cost of convolutions $\phi_{conv} = I^2r^2 c m_1 + \sum_{i=1}^{k-1} I^2r^2 m_{i}m_{i+1}$ and the cost of pooling $\phi_{pool} = \sum_{i=1}^k I^2p_im_i$. After the convolutional layers is an MLP of input size $I^2p_km_k$, whose features we also include. Our final cost model is thus:
$$c(x) = c_1 \phi_{conv} + c_2 \phi_{pool} +  c_3 \phi_{quad} + c_4 \phi_{linear} + c_5.$$

\begin{figure}[t]
    \centering
    \includegraphics[scale=0.5]{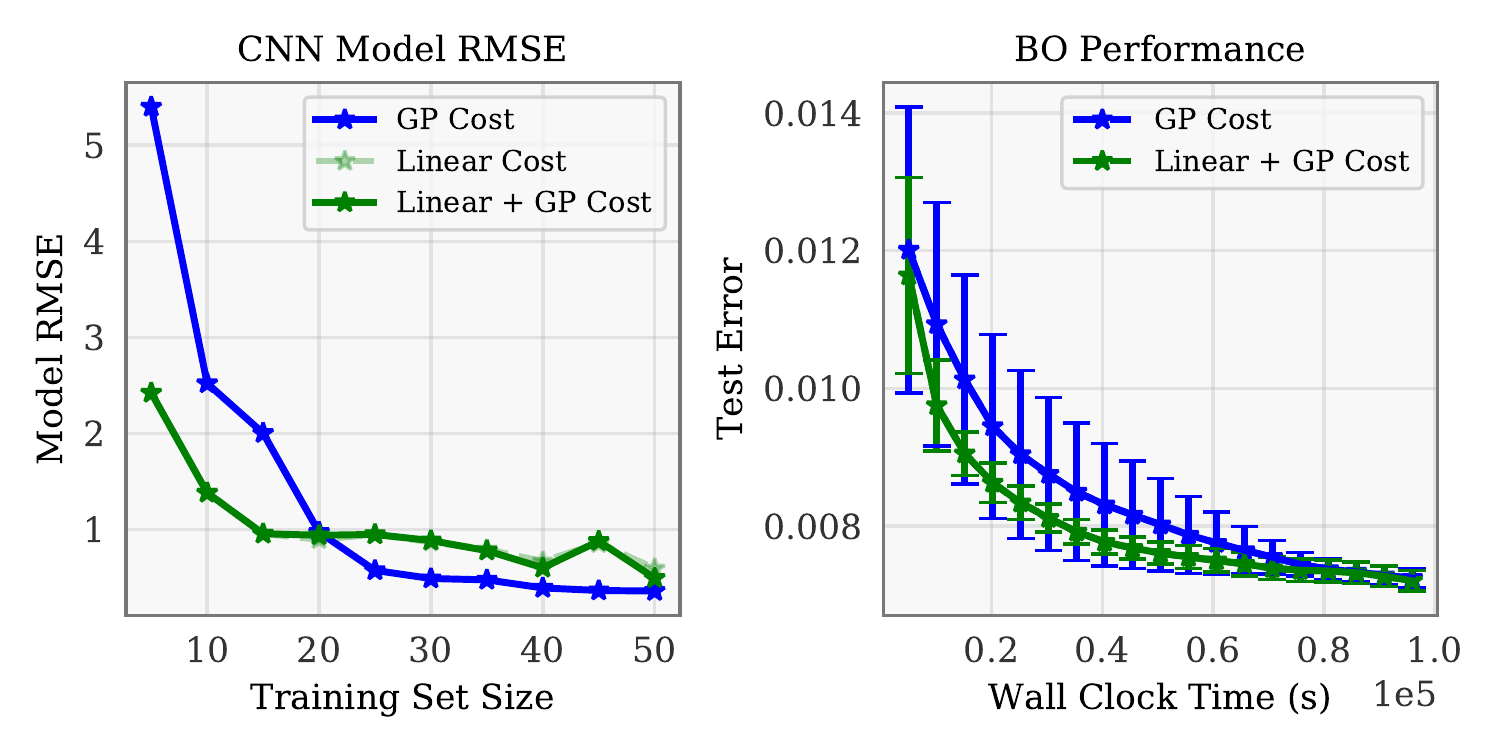}
    \caption{The low-variance CNN model had lower RMSE only in the limited data regime (iterations $<$ 20). Though it converges faster than the warped GP, both converge to the same optimum.}
    \label{fig:cnn_performance}
\end{figure}

We let $h=4$, $1 \leq r \leq 10$, $8 \leq m_i \leq 128$, $k = 2$, $8 \leq n_i \leq 128$, $b = 100$, $4 \leq e \leq 20$, and we add a single max pooling layer after the convolutional layers and a dropout rate of 0.25. We train on MNIST \cite{lecun1998gradient}, and compare accuracy between two cost models: a warped GP and a GP whose mean is the low-variance model. We train on timings consistent with EIpu evaluation points and compare classification error using 10,000 random hyperparameter configurations. CNNs proved more difficult to model than MLPs. Figure~\ref{fig:cnn_performance} shows that the low-variance model is only better than the GP for small training sets of size less than 20. This is likely because flops do not reflect the actual runtime of CNN training, which uses highly optimized libraries to perform convolutions. 

\section{Conclusion}\label{Conclusion}
How to best use a cost model to plan optimization when evaluation cost varies is a challenging question. EIpu, which normalizes the acquisition function by the cost model, is reasonable insofar as having the correct units, but performs poorly if the optimum is not cheap. We introduced CArBO, an algorithm adapting the \textit{early and cheap, late and expensive} strategy from prior work on grey-box, cost-aware BO to the black-box setting, in which no external information about cost is given. By combining a cost-effective initial design and cost-cooling, CArBO was shown to outperform EI and EIpu on an extensive set of real-world benchmarks, both in the sequential and batch setting. Additionally, we showed that the performance of cost-aware BO can further benefit from low-variance cost models that extrapolate well.
6
A number of directions for future work are open. Adapting CArBO's initial design and cost-cooling to other acquisition functions, such as predictive entropy search \cite{hernandez2014predictive} or max-value entropy search~\cite{wang17e}, is straightforward. Combining CArBO with multi-fidelity to learn fidelity parameters and their relationship to cost is also of interest. As we showed, building an accurate cost model is an important problem, and our flop-counting approach can certainly be built on. Finally, CArBO assumes fixed batch size, and allowing it to vary may boost performance further. This is likely of practical importance as BO is often run on clusters or cloud services.

\bibliography{bibliography}
\bibliographystyle{icml2020}

\newpage
\appendix
\section*{Supplementary material}

\section{Batch Bayesian Optimization}
In the batch setting, $b$ candidates $\xb^1, \dots, \xb^b$ are evaluated in parallel. Batch fantasizing extends EI to the batch setting by predicting multiple future evaluation trajectories, known as ``fantasies''. These fantasies are aggregated to identify a sequence of evaluation points, which is then evaluated in parallel. A batch can be obtained through sequential fantasizing as described Algorithm~\ref{algorithm_batch}.

\begin{algorithm}[h]
\caption{Batch Fantasizing with EI}
\label{algorithm_batch}
\begin{algorithmic}[1]
  \STATE {\bfseries Input:} batch $b$, fantasies $n_f$, data $\Xb,Y$
    \FOR{$i = 1, \dots, n_f$}
        \STATE $Y^{(i)} \gets \texttt{copy}(Y)$ 
    \ENDFOR
    \STATE $\xb_1 \gets \texttt{argmax}_{\xb \in \Omega} \; \text{EI}(\xb \;|\;  \Xb, Y ) $ 
    \FOR{$j = 2, \dots, b$}
        \STATE $\Xb \gets \Xb \cup \{\xb_j \}$
        \FOR{$i = 1, \dots, n_f$}
            \STATE $y_k^{(i)} \gets \texttt{sample\_posterior}(\xb \;|\;  \Xb, Y^{(i)})$ 
            \STATE $ Y^{(i)} \gets Y^{(i)} \cup \{ y^{(i)}\}$
        \ENDFOR
        \STATE $\xb_j \gets \texttt{argmax}_{\xb \in \Omega} \frac{1}{n_f} \sum_{i=1}^{n_f} \text{EI}(\xb \;|\;  \Xb, Y^{(i)})$
    \ENDFOR
    \STATE \textbf{return} $\Xb_b = \{ \xb_1, \dots, \xb_b\}$
\end{algorithmic}  
\end{algorithm}

Batch fantasizing using EI, without loss of generality, works as follows. First, let $\xb_1$ be the the argmax of EI, and sample $n_f$ values from the posterior at $\xb_1$. These $n_f$ ``fantasy'' values represent possible future evaluation trajectories. We maintain $n_f$ different GPs, and update them each with a different fantasy value (GP hyperparameters are kept constant). Second, we maximize a new acquisition that is the average of EI acquisition functions over each different fantasy. The argmax of this averaged acquisition is set to be $\xb_2$. We repeat these steps until a batch of $b$ points is obtained. In CArBO, we set $n_f$ to be 10 by default. 

\section{Cost Cooling}
\begin{figure}[t]
    \centering
    \includegraphics[scale=0.5]{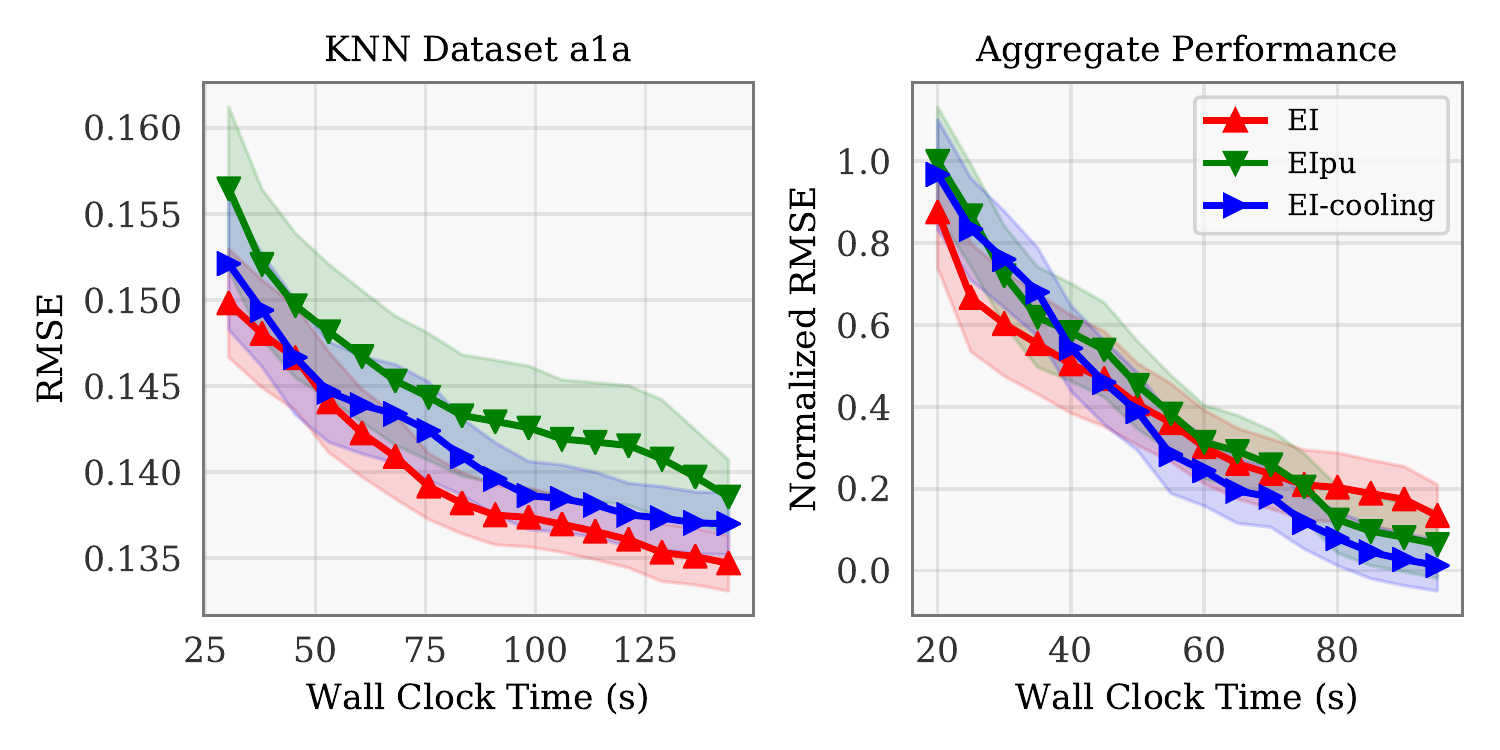}
    \caption{(\textit{Left}) EI-cool is run on the same KNN problem in Figure 2 (Section 2) of our paper. It outperforms EIpu but not EI. In general, EI-cool usually outperforms at least one of EIpu or EI, but not always both. (\textit{Right}) This results in more robust performance, as demonstrated by EI-cool's superior performance over EI and EIpu when aggregated over four KNN benchmarks. }
    \label{fig:knncostcooling}
\end{figure}

Cost-cooling deprecates the cost model as iteration proceeds. Assume that at the $k$th BO iteration, $\tau_k$ of the total budget $\tau$ has been used (at $k=0$, $\tau_k = \tau_{init}$). Cost-cooling, which we call EI-cool when using EI, is defined as:
\begin{equation}\label{ei-cool}
\text{EI-cool}(\xb) \coloneqq \frac{\EI(\xb)}{c(\xb)^\alpha} \;,\; \alpha = (\tau - \tau_k) / (\tau - \tau_{init}).
\end{equation}
As the parameter $\alpha$ decays from one to zero, EI-cool transitions from EIpu to EI. EI-cool is not meant to beat both EI and EIpu on every benchmark but can be rather thought of as a compromise between the two. We found it is not guaranteed to outperform both but usually outperforms \textit{at least one}. This is shown in Figure~\ref{fig:knncostcooling}, in which EI-cool is better than EIpu but worse than EI on a specific KNN problem, but performs better than both in the aggregate. 

\section{CArBO}
\begin{figure*}[t!]
    \centering
    \includegraphics[scale=0.65]{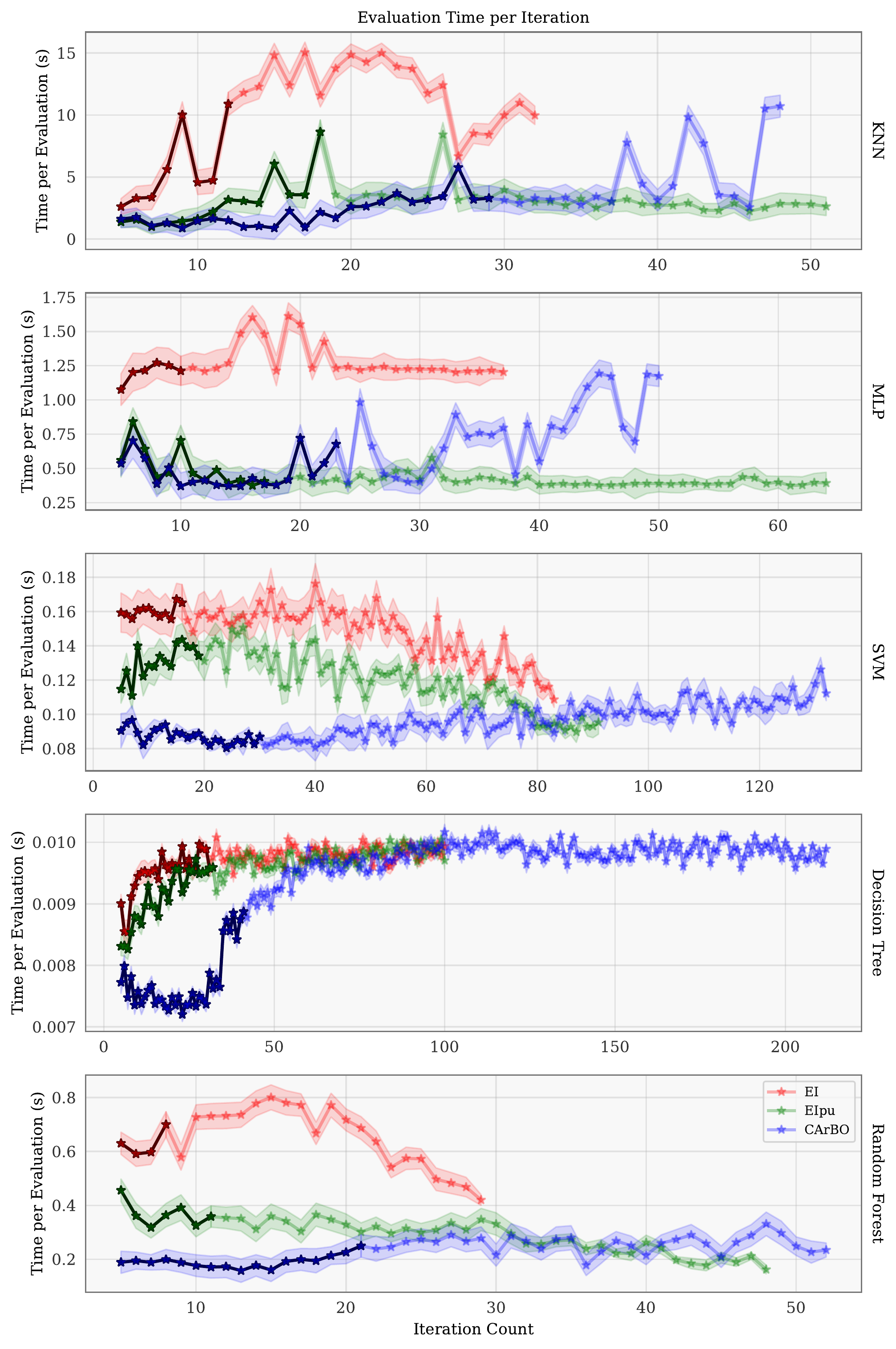}
    \caption{Evaluation times for KNN, MLP, SVM, Decision Tree, and Random Forest, on the a3a dataset. EI (blue) and EIpu (green) do not perform any explicit scheduling, and thus have widely varying evaluation times. CArBO clearly performs cheap evaluations before expensive ones, as seen in all five subplots. This enables it to converge faster than both EI and EIpu. } 
    \label{fig:opt_timing_all}
\end{figure*}

CArBO effectively schedules cheap evaluations before expensive ones, without the need for any additional information. This is a general trend evidenced by the subplots in Figure~\ref{fig:opt_timing_all}, which demonstrates that CArBO clearly schedules cheap evaluations before expensive ones. Note that CArBO performs better on all five of these problems. 

\end{document}

%% file: macros.tex
\def\xb{{\mathbf x}}

\def\Xb{{\mathbf X}}

\DeclareMathOperator*{\argmin}{arg\,min}

